\documentclass{article}
\usepackage{spconf,amsmath,graphicx}
\usepackage[utf8]{inputenc}
\usepackage[dvipsnames]{xcolor}

\title{Linguistic unit discovery from multi-modal inputs in unwritten languages: Summary of the ``Speaking Rosetta'' JSALT 2017 Workshop} %

\name{\begin{tabular}{c}
        Odette Scharenborg$^{1}$\sthanks{Corresponding author: O.Scharenborg@let.ru.nl},
        Laurent Besacier$^{2}$,
        Alan Black$^{3}$,
        Mark Hasegawa-Johnson$^{4}$,\\
        Florian Metze$^{3}$,
        Graham Neubig$^{3}$,
        Sebastian Stüker$^{5}$,
        Pierre Godard$^{6}$,
        Markus Müller$^{5}$,\\
        Lucas Ondel$^{8}$,
        Shruti Palaskar$^{3}$,
        Philip Arthur$^{3}$,
        Francesco Ciannella$^{3}$, 
        Mingxing Du$^{7}$,\\
        Elin Larsen$^{7}$, 
        Danny Merkx$^{1}$,
        Rachid Riad$^{7}$,
        Liming Wang$^{4}$,
        Emmanuel Dupoux$^{7}$
        \sthanks{The work reported here was started at JSALT 2017 in CMU, Pittsburgh, and was supported by JHU and CMU via grants from Google, Microsoft, Amazon, Facebook, Apple. This work used the Extreme Science and Engineering Discovery Environment (XSEDE), which is supported by NSF grant number OCI-1053575.  Specifically, it used the Bridges system, which is supported by NSF award number ACI-1445606, at the Pittsburgh Supercomputing Center (PSC). OS was partially supported by a Vidi-grant from NWO (276-89-003). PG was funded by the French ANR and the German DFG under grant ANR-14-CE35-0002 (BULB project). MD, EL, RR and ED were funded by the European Research Council (ERC-2011-AdG-295810 BOOTPHON), and ANR-10-LABX-0087 IEC and ANR-10-IDEX-0001-02 PSL*.
        }
        \end{tabular}}

\address{
$^{1}$ Radboud University, 
$^{2}$ LIG - Univ Grenoble Alpes (UGA),
$^{3}$ Carnegie Mellon University, 
\\
$^{4}$ University of Illinois, 
$^{5}$ Karlsruhe Institute of Technology, 
$^{6}$ LIMSI CNRS, 
\\
$^{7}$ ENS/CNRS/EHESS/INRIA, 
$^{8}$ Brno University.
}

\begin{document}
%
\maketitle
\begin{abstract}
We summarize the accomplishments of a multi-disciplinary workshop exploring the computational and scientific issues surrounding the discovery of linguistic units (subwords and words) in a language without orthography. We study the replacement of orthographic transcriptions by images and/or translated text in a well-resourced language to help unsupervised discovery from raw speech.
\end{abstract}
\begin{keywords}
unwritten languages, multi-modal data, unsupervised unit discovery, image retrieval, machine translation.
\end{keywords}

\vspace{-0.4cm}
\section{Introduction}
To develop speech and language technology (SLT) large amounts of annotated data are required. However, for many languages in the world, not enough speech data is available, or these lack the annotations needed to train an ASR system \cite{addaBulbSLTU2016}. Moreover, an estimated half of the human languages do not have an orthography, and many others do not use it in a consistent fashion. This represents millions of potential users that as yet cannot be served by speech technologies. As any human 4-year-old demonstrates, however, it is theoretically possible to learn a language communication system before learning to read and write, from raw sensory signals and with only limited human supervision. 

Recently, different approaches have been proposed to build ASR systems for such low-resource languages. One strand of research focuses on discovering the linguistic units of the low-resource language from the raw speech data, while assuming no other information about the language is available, and using these to build ASR systems (zero resource approach; e.g., \cite{jansen_2013,ondel2016,varadarajan2008,park2008,zhang2010}). Another strand of research focuses on building ASR systems using speech data from multiple languages, thus trying to create universal or cross-lingual ASR systems \cite{schultz2001,loof2009,vesely2012,xu2015}. Children though, when learning a language, also have information besides the auditory input available, primarily in the visual modality. This has led to a new strand of research which uses visual information, from images, to discover word-like units from the speech signal using speech-image associations \cite{harwarth2015,chrupala2017,harwarth2016}. The ``Speaking Rosetta'' project at the 2017 Frederick Jelinek Memorial Summer Workshop, which took place at Carnegie Mellon University, Pittsburgh, pushed this idea further by using multi-modal datasets that not only include images, but also include translations in a high-resource language. This is an interesting extension as parallel data between speech from an unwritten language and translations of that speech signal in another language can easily be collected \cite{blachon2016}.
\vspace{0.2cm}
\\
This paper summarizes the accomplishments of the multidisciplinary ``Speaking Rosetta'' workshop which explored the computational and scientific issues surrounding the discovery of linguistic units (subwords and words) in a language without orthography, through replacing the orthographic transcriptions typically used for training an ASR system by images and/or translations in a well-resourced language. The focus of the project was on discovering intermediate symbolic units and investigating their role in building SLT systems. We concentrated on 4 tasks: two with symbolic units (unit discovery and speech synthesis) and two end-to-end tasks without the need for explicit symbolic units (speech2image and speech2translation).

\vspace{-0.4cm}
\section{Overview of ``Speaking Rosetta''}
\begin{figure}
    \centering
    \includegraphics[width=7cm]{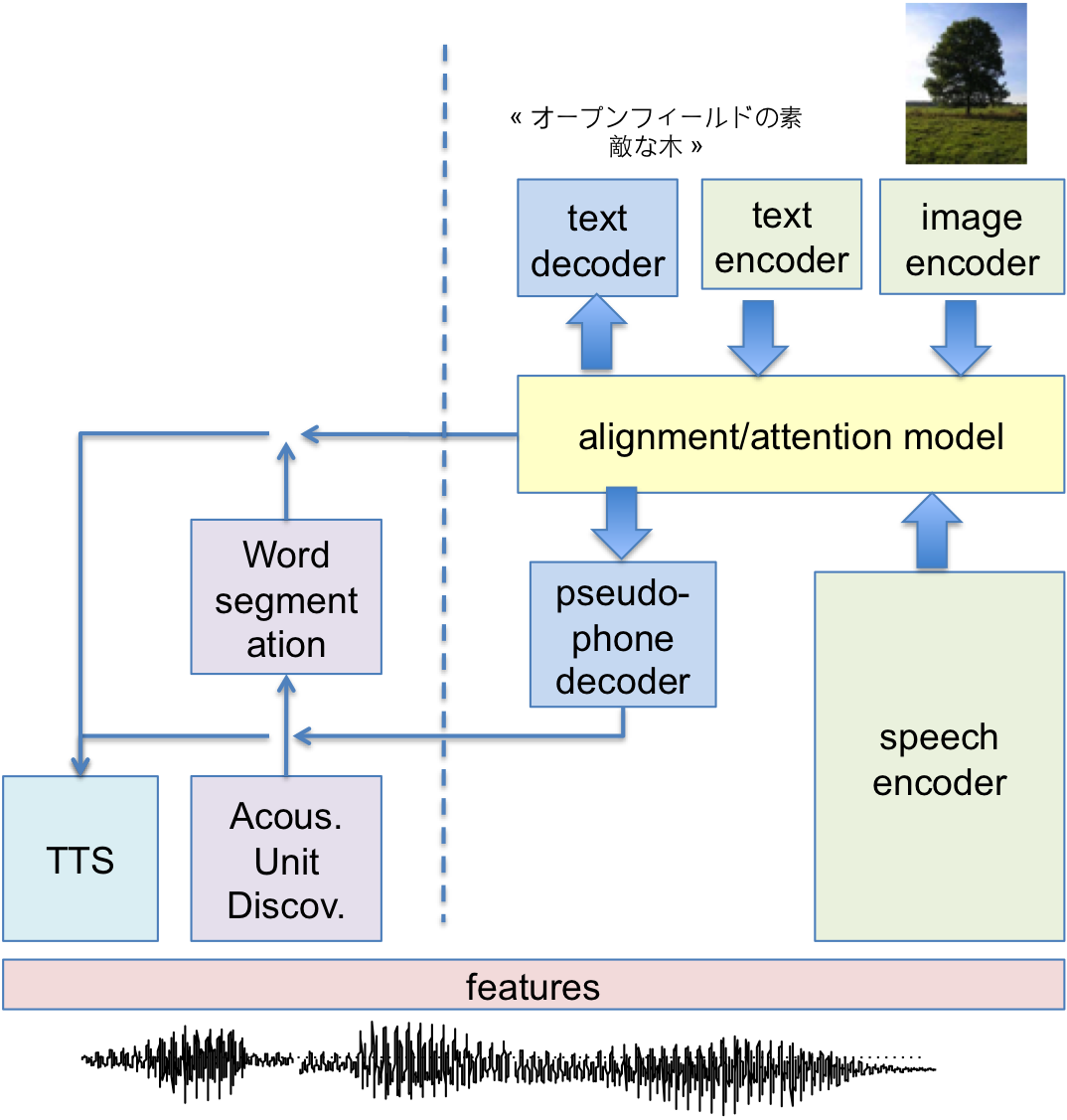}
    \caption{Functional blocks of the ``Speaking Rosetta'' project}
    \label{fig:outlineproject}
\end{figure}

Figure \ref{fig:outlineproject} shows a visual representation of the end-to-end systems, and structure, of the Rosetta project. The unit discovery strand (see Section 3.1) focused on discovering 'acoustic units' in the form of articulatory features or (pseudo) phones from raw speech. These acoustic units were used to build speech synthesis systems (Section 3.2), to transform the speech input into symbolic units (pseudo words or pseudo phones) and these units were used for several end-to-end tasks. The end-to-end tasks (see Section 4) used an encoder-decoder framework to translate speech or retrieve images from speech. Alignment/attention models were taken advantage of. Two of these end-to-end tasks are highlighted below: speech to translation and speech to image retrieval. 

\vspace{-0.4cm}
\subsection{Databases}
Five multi- and unimodal databases were used. The \textbf{Mboshi} (Bantu language spoken in Congo-Brazzavile) corpus\footnote{The dataset will be made available for free by ELRA; its current version is online at: https://github.com/besacier/mboshi-french-parallel-corpus} consists of 5k speech utterances (approximately 4 hours of speech) in Mboshi aligned to French text. The data set also contains linguists’ transcriptions in Mboshi in the form of a non-standard graphemic form close to the language phonology  \cite{addaBulbSLTU2016, mboshi-arxiv}. 

The \textbf{FlickR-real speech} database is a tri-modal (speech, translated text, images) corpus. The FlickR corpus contains 5 different natural language text captions (obtained using Amazon Mechanical Turk; AMT) for each of 8000 images captured from the FlickR photo sharing website. AMT was also used by 
\cite{Harwath2015} to obtain 40K spoken versions of the captions. We augmented this corpus by adding Japanese translations (Google MT) for all 40K captions, as well as Japanese tokenization.

\textbf{SPEECH-COCO-synthetic}
\cite{ds80,Havard2017} is an augmentation of MSCOCO \cite{MSCOCO} which consists of 123,287 images with five different descriptions per image. We generated speech captions using text-to-speech (TTS) synthesis resulting in 616,767 spoken captions (more than 600h) paired with images. Disfluencies and speed perturbation were added to the signal in order to make it sound more natural. 

The \textbf{How-To dataset} is an English open domain instructional videos (uploaded by users with personal video recorders) dataset of about 480 hours of speech. Each video is broken down into short utterances of about 8-10 seconds each. Transcriptions consist of summaries of what was spoken. The cleanest 45 hours out of the 480 hours were used.

From the \textbf{Spoken Dutch Corpus} (CGN, \cite{oostdijk2002}), 64 hours of read speech were used.
\vspace{-0.4cm}
\subsection{Evaluation}
The two types of task, i.e., linguistic unit discovery and end-to-end, were evaluated using a battery of tests which include qualitative measures, e.g., the MCD (see Section 3.2 \cite{Black2006}) and the ABX task (which compares the similarity between discovered units and ground truth labels or between different types of acoustic features; \cite{schatz2013,schatz2014}) for the evaluation of the discovered units, and quantitative measures, such as BLEU score, error rates, and word discovery metrics (see for more details \cite{ludusan2014,zrc2017}).  

\vspace{-0.4cm}
\subsection{XNMT Toolkit}
The end-to-end systems used during the project were built using the neural machine translation toolkit XNMT~\cite{xnmt},
which was greatly improved during the course of this project. XNMT is a sequence-to-sequence neural network toolkit which reads in a sequence of (variable-length) inputs, and then generates a different sequence of (variable-length) output. It consists of a library of standard components. The library is designed so that existing components can be easily re-arranged to run new experiments, and new components can be easily added. Available components are categorized as embedders (e.g., onehot, linear, and continuous vector embedders), encoders (e.g., CNN, LSTM, and pyramidal LSTM encoders), attention models (e.g., dot product, bilinear, and MLP attention models), decoders (e.g., a RNN decoder applied to the state vector of the encoder), and error metrics (e.g., BLEU, cross-entropy, word error rate).

\vspace{-0.4cm}
\section{Tasks with symbolic units}

\subsection{Unit discovery}
Three different unit discovery systems were implemented that used out-of-domain languages to help unit discovery through (almost) zero-shot adaptation.

In the \textbf{unsupervised phoneme discovery - Bayesian acoustic unit discovery (AUD)} approach, pseudo-phones were generated from the AUD system of \cite{ondel2016} with two major modifications. First, the truncated Dirichlet process of \cite{ondel2016} was replaced by a symmetric Dirichlet distribution, which provides a good and yet simple approximation of the Dirichlet Process \cite{Kurihara2007}. Second, to cope with larger databases, the Variational Bayes Inference algorithm originally used in \cite{ondel2016} was replaced with the faster Stochastic Variational Bayes Inference algorithm. Experiments showed that these modifications, while considerably speeding up the training, yielded negligible drop in accuracy. Also, an extension of this model was explored: the AUD model was embedded into a Variational Auto-Encoder leading to a specific case of the recently developed Structured Variational Auto-Encoder model \cite{Johnson2016}\footnote{The source code of both AUD models is available via https://github.com/amdtkdev/amdtk}. 

The \textbf{universal articulatory features and phoneme inventory discovery} approach aimed at deriving phone-like units using the setup presented in \cite{essv2017mueller}. It consists of three steps: 1) Detection of pseudo-phone boundaries 2) Extraction of language-universal articulatory features (AFs) for each segment. 3) Clustering of the segments based on the extracted AFs. Seven articulatory feature detectors using different network architectures were trained using data from multiple source languages, and evaluated cross-lingually.  Results indicated that the LSTM-based feature extractors showed an improved multilingual performance compared to \cite{essv2017mueller}, but they did not perform as good as their feed-forward neural network based counterparts crosslingually. Using k-means, segments were clustered based on the extracted AFs of each segment. Estimating the number of classes k is an open question for future research.

The \textbf{cross-language definition of units} approach \cite{Scharenborg_Casablanca} uses linguistic knowledge of the low-resource language and a semi-supervised training paradigm to build an ASR system for a low-resource language through the adaptation of an ASR system of a high-resource language. Crucially, phones that are present in the low-resource language but not in the high-resource language need to be created. This is done through a linear extrapolation between existing acoustic units in the high-resource ASR system's soft-max layer after which the acoustic units are iteratively retrained using all utterances or only those that have the best score according to four different criteria: ASR score, the MCD score from a TTS system (see Section 3.2), translated text retrieval score, and their combination. The experiments showed that in order to train acoustic units using self-labelled data, training utterances are needed that capture multiple aspects of the speech signal.

\vspace{-0.4cm}
\subsection{``TTS without T''}
Text-to-speech (TTS) technology was used to generate speech from unit sequences, and to evaluate the quality of the discovered unit inventories. Since this project concerns languages without orthography, TTS systems need to be built using discovered units rather than text (dubbed ``TTS without T''). The TTS system used is Clustergen \cite{Black2006}. Clustergen works well with small corpora because it treats each frame of the training corpus as a training example, rather than each segment. This makes it suitable for our low-resource scenario. The input to Clustergen is a waveform file plus symbolic sequences of “phones”; the output is a simple synthesizer and a Melcepstral distortion measure (MCD) \cite{toda2004} on held out data. MCD measures the average distance between the log-spectra of the synthetic and natural utterances, and has been demonstrated to be an extremely sensitive measure of the perceived naturalness of speech utterances, e.g., an MCD difference between two synthesis algorithms of 0.3 (on the same test corpus) is usually perceptible by human listeners as a significant difference in perceived naturalness \cite{Black2006}. 

TTS was used to generate speech in two tasks. The first task is a new speech technology task, which we call 
\textbf{image2speech} \cite{hasegawa-johnson2017}. Image2speech is similar to automatic image captioning, but can reach people whose language does not have a natural or easily used written form. The image2speech pipeline consists of a VGG16 visual object recognizer which converts each image into a sequence of feature vectors. XNMT accepts image feature vectors as inputs, and generates speech units as output, which were then sent to the TTS. Four types of intermediate speech units were tested: 1) L1-words and 2) L1-phones (generated using a same-language ASR, which provides an upper bound performance); 3) L2-phones from the cross-language definition of units approach and 4) pseudo-phones generated using AUD (see Section 3.1 for both). Results showed that the image2speech system is able to generate a phone string that is composed entirely of intelligible words, sequenced in an intelligible and semantically reasonable sentence.

In the second task, a proof-of-concept \textbf{foreign-text-2-speech} end-to-end system was build using XNMT which translates French words (text) into Mboshi phones which were either (1) true phones (2) or pseudo-phones obtained via AUD (see Section 3.1). These phone sequences were then sent to the TTS system. On a development set of 514 utterances BLEU4 scores at the character level were of 31.95\% with true phones and 8.32\% with pseudo-phones\footnote{TTS speech samples are available via https://github.com/JSALT-Rosetta/Illustrations/blob/master/TTS/mboshi/}.

\vspace{-0.4cm}
\subsection{Speech and image to text (and summarization)}
The speech-and-image2text system uses multi-modal information consisting of speech and videos to improve standard (supervised) ASR (this approach is thus also useful for high-resource languages). From the videos, object and scene features are extracted and used to adapt a sequence-to-sequence model (using the Pyramidal encoder by \cite{LAS}) to the visual features. Results showed that adding the visual features helps the model convergence and guides the training in the earlier epochs, compared to an HMM-DNN model. The sequence-to-sequence model is able to jointly learn the audio visual features, the acoustic and language models, requires no extra preprocessing for noisy data, does not require precomputed alignments, and is efficient even with long utterances. 

\vspace{-0.4cm}
\section{End-to-end tasks}

\subsection{Speech-to-translation}

End-to-End speech translation, i.e., translation from raw speech without any intermediate transcription \cite{berard2016listen,weiss2017sequence}, is attractive for language documentation, which often uses corpora made of audio recordings aligned with their translation in another language (no transcript in the source language) \cite{addaBulbSLTU2016,blachon2016}. Here, XNMT was used to build end-to-end speech translations systems on FlickR (English-to-Japanese) and Mboshi-to-French. The obtained BLEU4 scores at the character level were 30.99\%  and 22.36\% on the development sets of FlickR and Mboshi, respectively. Although these results are rather low for a pure translation task, these systems show that end-to-end models are able to encode some regularities in the speech signal in order to decode predictable sequences of characters in a target language.

Secondly, an attention-based Neural Machine Translation (NMT) model \cite{zanon2017} was trained between phones in Mboshi and text in French, while soft-alignment probability matrices generated by the attention mechanism, were extracted. These alignments were post-processed to segment a sequence of symbols in Mboshi into words. While \cite{zanon2017} applied their method to true phones (gold phonemes transcribed by linguists), here segmentation through attention from a pseudo-phone sequence obtained using AUD (see Section 3.1) was investigated. 
Table \ref{tab:attention} shows that the word boundary detection results of the attention-based system outperformed those of a pure speech-based baseline which used pair-matching using locally sensitive hashing applied to PLP features and then grouped pairs using graph clustering \cite{aren}. Moreover, a reverse model (French-Mboshi) slightly improved word segmentation compared to (Mboshi-French). Implementation of a bilingual loss is probably an interesting future work. 

\begin{table}
    \centering
    \begin{tabular}{|l|c|c|c|c}
         \hline
         \textbf{System} & \textbf{Prec} & \textbf{Recall} & \textbf{F} \\
         \hline
         Segmental DTW Baseline \cite{aren} & 31.9 & 13.8 & 19.3 \\
        Attention (fr-mb)      & 36.5 & 46.1 & 40.7 \\
        Attention (mb-fr)      & 36.3 & 46.6 & 40.8 \\
        \hline
    \end{tabular}
    \caption{Speech-to-translation: Word boundary detection results (Mboshi5k corpus) from pseudo phones}
    \label{tab:attention}
\end{table}

\vspace{-0.4cm}
\subsection{Speech-to-Image}
Speech-to-image is a relatively new task \cite{harwarth2015,chrupala2017,harwarth2016}. A speech-to-image system learns to map images and speech to the same embedding space, and retrieves an image using spoken captions. While doing so, it uses multi-modal input to discover speech units in an unsupervised manner, similar to how children acquire their first language. Our speech-to-image system (based on the implementation of \cite{harwarth2016}) was implemented using XNMT. Four types of acoustic features were compared: Mel-frequency Filterbanks (baseline, similar to \cite{Harwath2015} but with added speaker-dependent mean-variance normalization on the features before zero-padding/truncation), the pseudo-phones generated by the AUD system \cite{ondel2016} (which were downsampled by a factor of 9 along the phone dimension to fit the input of the DNN), Multilingual Bottleneck features (MBN), and Cochleagram Features generated by the Resonant Tectorial Model developed by \cite{AllenSen1999}.

Table \ref{tab:featcomp} shows the results for the four features evaluated with Recall@N. The MBN feature is superior to all other acoustic features, and shows over 1 percent improvement on the Filterbank baseline for the recall@10 score. 

\begin{table}
    \centering
    \begin{tabular}{|l|c|c|c|}
        \hline
        \textbf{Feature type} & \textbf{R@1} & \textbf{R@5} & \textbf{R@10} \\
         \hline
         Mel-filterbank & 0.0096 & 0.047 & 0.0856 \\
         Multiling. Bottleneck & \textbf{0.013} & \textbf{0.053} & \textbf{0.0994} \\  
         AUD (one epoch) & 0.0012 & 0.0044 & 0.0112 \\
         Cochleagram & 0.0008 & 0.005 & 0.0104 \\
         \hline
    \end{tabular}
    \caption{Speech-to-image retrieval results (Recall@N) for the tested input speech features}
    \label{tab:featcomp}
\end{table}

\vspace{-0.4cm}
\section{Concluding remarks}
The ``Speaking Rosetta'' JSALT 2017 project laid the foundation for a new research area ``Unsupervised multi-modal language acquisition''. It showed that it is possible to build useful SLT systems without any textual resources in the language for which the SLT is built, in a way that is similar to that of how infants learn a language. 1) The ``Speaking Rosetta'' project showed that zero-shot adaptation, i.e., unsupervised learning of speech units, is possible, and can be improved by using information extracted from well-resourced languages. The discovered units are meaningful as shown by their usefulness in upstream tasks such as word discovery,  image retrieval, and speech translation tasks. We have presented the first attempt to discover spoken term from speech using an attention matrix; the performance of this approach is better than all the baselines evaluated in the same conditions. 2) TTS has proven to be a useful tool in the evaluation of discovered units of different types, and can be used to evaluate how well a particular set of units correlates with acoustic features. “Units” we have tested include articulatory features, AUDs, and cross-language adapted phones. 3) The unit-discovery and the end-to-end systems were successfully combined into several working proof-of-concept end-to-end demos. 4) We showed that audio and video information can be fused to improve speech summarization without going through text. 5) Finally, a pipeline of metrics as well as dedicated datasets were created to fuel reproducible researches in this new emerging domain.

\newpage

{
\small
\baselineskip 5.0 mm
\renewcommand\refname{}
\bibliographystyle{IEEEbib}
\bibliography{main}

\begin{thebibliography}{10}

\bibitem{addaBulbSLTU2016}
G.~Adda et~al.,
\newblock ``Breaking the unwritten kanguage barrier: The {Bulb} project,''
\newblock in {\em Proceedings of SLTU}, Yogyakarta, Indonesia, 2016.

\bibitem{jansen_2013}
A.~Jansen et~al.,
\newblock ``A summary of the 2012 {JH} {CLSP} {Workshop} on zero resource
  speech technologies and models of early language acquisition,''
\newblock in {\em Proceedings of ICASSP}, 2013.

\bibitem{ondel2016}
L.~Ondel, L.~Burget, and J.~{\v{C}}ernock{\'{y}},
\newblock ``Variational inference for acoustic unit discovery,''
\newblock in {\em Procedia Computer Science}, 2016, pp. 80--86.

\bibitem{varadarajan2008}
B.~Varadarajan, S.~Khudanpur, and E.~Dupoux,
\newblock ``Unsupervised learning of acoustic sub-word units,''
\newblock in {\em Proceedings of ACL on Human Language Technologies: Short
  Papers}, 2008, pp. 165--168.

\bibitem{park2008}
A.~S. Park and J.~R. Glass,
\newblock ``Unsupervised {Pattern} {Discovery} in {Speech},''
\newblock {\em IEEE Transactions on Audio, Speech, and Language Processing},
  vol. 16, no. 1, pp. 186--197, 2008.

\bibitem{zhang2010}
Y.~Zhang and J.~R. Glass,
\newblock ``Towards multi-speaker unsupervised speech pattern discovery,''
\newblock in {\em Proceeding of ICASSP}, 2010, pp. 4366--4369.

\bibitem{schultz2001}
A.~W. Tanja~Schultz,
\newblock ``Experiments on cross-language acoustic modelling,''
\newblock in {\em Proceedings of Interspeech}, 2001.

\bibitem{loof2009}
J.~Lööf, C.~Gollan, and H.~Ney,
\newblock ``Cross-language bootstrapping for unsupervised acoustic model
  training: rapid development of a polish speech recognition system,''
\newblock in {\em Proceedings of Interspeech}, 2009.

\bibitem{vesely2012}
K.~Vesely, M.~Karafiát, F.~Grezl, M.~Janda, and E.~Egorova,
\newblock ``The language-independent bottleneck features,''
\newblock in {\em Proceedings of SLT}, 2012.

\bibitem{xu2015}
H.~Xu, V.~Do, X.~Xiao, and E.~Chng,
\newblock ``A comparative study of bnf and dnn multilingual training on
  cross-lingual low-resource speech recognition,''
\newblock in {\em Proceedings of Interspeech}, 2015, pp. 2132--2136.

\bibitem{harwarth2015}
D.~Harwarth and J.~Glass,
\newblock ``Deep multimodal semantic embeddings for speech and images,''
\newblock in {\em Proceedings ASRU}, 2015, pp. 237--244.

\bibitem{chrupala2017}
G.~Chrupała, L.~Gelderloos, and A.~Alishahi,
\newblock ``Representations of language in a model of visually grounded speech
  signal,''
\newblock in {\em Proceedings of ASRU}, 2017.

\bibitem{harwarth2016}
D.~Harwath, A.~Torralba, and J.~Glass,
\newblock ``Unsupervised learning of spoken language with visual context,''
\newblock in {\em Advances in Neural Information Processing System}, 2016, pp.
  1858--1866.

\bibitem{blachon2016}
D.~Blachon, E.~Gauthier, L.~Besacier, G.-N. Kouarata, M.~Adda-Decker, and
  A.~Rialland,
\newblock ``Parallel speech collection for under-resourced language studies
  using the {LIG-Aikuma} mobile device app,''
\newblock in {\em Proceedings of SLTU}, Yogyakarta, Indonesia, May 2016.

\bibitem{mboshi-arxiv}
P.~Godard et~al.,
\newblock ``A very low resource language speech corpus for computational
  language documentation experiments,''
\newblock in {\em arXiv:1710.03501}, 2017.

\bibitem{Harwath2015}
D.~Harwath and J.~Glass,
\newblock ``Deep multimodal semantic embeddings for speech and images,''
\newblock in {\em Proceedings of ASRU}, Scottsdale, Arizona, USA, 2015, pp.
  237--244.

\bibitem{ds80}
L.~Besacier,
\newblock ``Speech-coco,''
  https://persyval-platform.univ-grenoble-alpes.fr/DS80/detaildataset.

\bibitem{Havard2017}
W.~Havard, L.~Besacier, and O.~Rosec,
\newblock ``Speech-coco: 600k visually grounded spoken captions aligned to
  mscoco data set,''
\newblock in {\em International Workshop on Grounding Language Understanding
  (GLU), Satellite of Interspeech 2017}, 2017.

\bibitem{MSCOCO}
T.-Y. Lin, M.~Maire, S.~Belongie, J.~Hays, P.~Perona, D.~Ramanan,
  P.~Doll{\'a}r, and C.~L. Zitnick,
\newblock ``Microsoft coco: Common objects in context,''
\newblock in {\em European Conference on Computer Vision (ECCV)}, Z{\"u}rich,
  2014,
\newblock Oral.

\bibitem{oostdijk2002}
N.~Oostdijk, W.~Goedertier, F.~V. Eynde, L.~Boves, J.-P. Martens, M.~Moortgat,
  and H.~Baayen,
\newblock ``Experiences from the spoken dutch corpus project,''
\newblock in {\em Proceedings of LREC, Las Palmas de Gran Canaria}, 2002, pp.
  340--347.

\bibitem{Black2006}
A.~W. Black,
\newblock ``{CLUSTERGEN}: A statistical parametric speech synthesizer using
  trajectory modeling,''
\newblock in {\em Proceedings of ICSLP}, 2006, pp. 1762--1765.

\bibitem{schatz2013}
T.~Schatz, V.~Peddinti, F.~Bach, A.~Jansen, H.~Hermansky, and E.~Dupoux,
\newblock ``Evaluating speech features with the {Minimal}-{Pair} {ABX} task
  ({I}): {Analysis} of the classical {MFC}/{PLP} pipeline,''
\newblock in {\em Proceedings of Interspeech}, 2013.

\bibitem{schatz2014}
T.~Schatz, V.~Peddinti, X.-N. Cao, F.~Bach, H.~Hermansky, and E.~Dupoux,
\newblock ``Evaluating speech features with the {Minimal}-{Pair} {ABX} task
  ({II}): {Resistance} to noise,''
\newblock in {\em Proceedings of Interspeech}, 2014.

\bibitem{ludusan2014}
B.~Ludusan, M.~Versteegh, A.~Jansen, G.~Gravier, X.-N. Cao, M.~Johnson, and
  E.~Dupoux,
\newblock ``Bridging the gap between speech technology and natural language
  processing: an evaluation toolbox for term discovery systems,''
\newblock in {\em Proceedings of {LREC}}, 2014.

\bibitem{zrc2017}
E.~Dunbar, X.~Nga~Cao, J.~Benjumea, J.~Karadayi, M.~Bernard, L.~Besacier,
  X.~Anguera, and E.~Dupoux,
\newblock ``The zero resource speech challenge 2017,''
\newblock in {\em Proceedings of ASRU}, 2017.

\bibitem{xnmt}
G.~Neubig,
\newblock ``Xnmt,'' https://github.com/neulab/xnmt/.

\bibitem{Kurihara2007}
K.~Kurihara, M.~Welling, and Y.~W. Teh,
\newblock ``Collapsed variational {D}irichlet process mixture models,''
\newblock in {\em Proceedings of the International Joint Conference on
  Artificial Intelligence}, 2007, vol.~20.

\bibitem{Johnson2016}
M.~J. Johnson, D.~Duvenaud, A.~B. Wiltschko, S.~R. Datta, and R.~P. Adams,
\newblock ``Composing graphical models with neural networks for structured
  representations and fast inference,''
\newblock in {\em Neural Information Processing Systems}, 2016.

\bibitem{essv2017mueller}
M.~M{\"u}ller, J.~Franke, S.~St{\"u}ker, and A.~Waibel,
\newblock ``Improving phoneme set discovery for documenting unwritten
  languages,''
\newblock {\em Elektronische Sprachsignalverarbeitung (ESSV) 2017}, 2017.

\bibitem{Scharenborg_Casablanca}
O.~Scharenborg, F.~Ciannella, S.~Palaskar, A.~Black, F.~Metze, L.~Ondel, and
  M.~Hasegawa-Johnson,
\newblock ``Building an asr system for a low-resource language through the
  adaptation of a high-resource language asr system: Preliminary results,''
\newblock in {\em Proceedings of ICNLSSP, Casablanca, Morocco}, 2017.

\bibitem{toda2004}
T.~Toda, A.~W. Black, and K.~Tokuda,
\newblock ``Mapping from articulatory movements to vocal tract spectrum with
  gaussian mixture model for articulatory speech synthesis,''
\newblock in {\em Proceedings of SSW5, Pittsburgh, PA}, 2004, pp. 31--36.

\bibitem{hasegawa-johnson2017}
M.~Hasegawa-Johnson, A.~Black, L.~Ondel, O.~Scharenborg, and F.~Ciannella,
\newblock ``Image2speech: Automatically generating audio descriptions of
  images,''
\newblock in {\em Proceedings of ICNLSSP, Casablanca, Morocco}, 2017.

\bibitem{LAS}
W.~Chan, N.~Jaitly, Q.~V. Le, and O.~Vinyals,
\newblock ``Listen, attend and spell,''
\newblock {\em arXiv preprint arXiv:1508.01211}, 2015.

\bibitem{berard2016listen}
A.~B{\'e}rard, O.~Pietquin, C.~Servan, and L.~Besacier,
\newblock ``Listen and translate: A proof of concept for end-to-end
  speech-to-text translation,''
\newblock in {\em NIPS workshop on End-to-end Learning for Speech and Audio
  Processing}, 2016.

\bibitem{weiss2017sequence}
R.~J. Weiss, J.~Chorowski, N.~Jaitly, Y.~Wu, and Z.~Chen,
\newblock ``Sequence-to-sequence models can directly transcribe foreign
  speech,''
\newblock {\em arXiv preprint arXiv:1703.08581}, 2017.

\bibitem{zanon2017}
M.~Zanon~Boito, A.~Berard, A.~Villavicencio, and L.~Besacier,
\newblock ``Unwritten languages demand attention too! word discovery with
  encoder-decoder models,''
\newblock in {\em Proceedings of ASRU}, 2017.

\bibitem{aren}
A.~Jansen and B.~Van~Durme,
\newblock ``Efficient spoken term discovery using randomized algorithms,''
\newblock in {\em Proceedings of ASRU}, 2011, pp. 401--406.

\bibitem{AllenSen1999}
J.~Allen and D.~Sen,
\newblock ``Is tectorial membrane filtering required to explain two tone
  suppression and the upward spread of masking?,''
\newblock in {\em Mechanics of Hearing}, 1999, p. 451.

\end{thebibliography}
}

\end{document}